\documentclass[
leqno
]{ceurart}

\usepackage{listings}
\usepackage{amssymb}
\usepackage{amsmath}
\usepackage{amsthm}
\usepackage{amsfonts}
\usepackage{stmaryrd}

\begin{document}

\copyrightyear{2024}
\copyrightclause{Copyright for this paper by its authors.
  Use permitted under Creative Commons License Attribution 4.0
  International (CC BY 4.0).}

\conference{The 23rd International Semantic Web Conference, November 11--15, 2025, Baltimore, MD}

\title{A Benchmark for the Detection of Metalinguistic Disagreements between LLMs and Knowledge Graphs}

% \tnotetext[1]{You can use this document as the template for preparing your
%   publication. We recommend using the latest version of the ceurart style.}

\author[1]{Bradley P. Allen}[%
orcid=0000-0003-0216-3930,
email=b.p.allen@uva.nl,
url=https://www.bradleypallen.org/,
]
\cormark[1]
\address[1]{University of Amsterdam, Science Park 900, 1098 XH Amsterdam, The Netherlands}

\author[1]{Paul T. Groth}[%
orcid=0000-0003-0183-6910,
email=p.t.groth@uva.nl,
url=https://pgroth.com/,
]

\cortext[1]{Corresponding author.}

\begin{abstract}
Evaluating large language models (LLMs) for tasks like fact extraction in support of knowledge graph construction frequently involves computing accuracy metrics using a ground truth benchmark based on a knowledge graph (KG). These evaluations assume that errors represent factual disagreements. However, human discourse frequently features \textit{metalinguistic disagreement}, where agents differ not on facts but on the meaning of the language used to express them. Given the complexity of natural language processing and generation using LLMs, we ask: do metalinguistic disagreements occur between LLMs and KGs? Based on an investigation using the T-REx knowledge alignment dataset, we hypothesize that metalinguistic disagreement does in fact occur between LLMs and KGs, with potential relevance for the practice of knowledge graph engineering. We propose a benchmark for evaluating the detection of factual and metalinguistic disagreements between LLMs and KGs. An initial proof of concept of such a benchmark is available on Github.
\end{abstract}



\maketitle

\section{Introduction}
\label{sec:intro}

Recent years have seen a surge of interest in the use of LLMs for purposes of knowledge engineering \cite{allen_et_al:TGDK.1.1.3}. LLMs are being used to perform text classification, sentiment analysis, and natural language inference, exploiting next-token prediction to generate text that can be transformed into the type of symbolic outputs normally produced in these tasks \cite{pretrainpromptpredict2023}. Increasing emphasis is being placed on the use of LLMs in knowledge graph construction \cite{koutsiana2024knowledge}. The results have been encouraging, but a major concern that has emerged is the impact of hallucination, which is defined as the presence of factually incorrect or unjustified assertions in the output of LLMs \cite{ji2023survey,huang2023survey}. Benchmarks such as SHROOM \cite{mickus-etal-2024-semeval} and WildHallucinations \cite{zhao2024wildhallucinations} have been developed to evaluate the ability to detect hallucination when it occurs in LLM output.

A number of mechanisms have been proposed to mitigate hallucination in LLMs through the use of knowledge from a variety of sources, including natural language text, KGs, and rules, to ground \cite{harnad2024language} an LLM. Retrieval-augmented generation (RAG) is a specific version of this approach that has attracted a great deal of interest, particularly in the context of commercial applications \cite{gao2023retrieval}. Such {\em knowledge-enhanced LLMs} \cite{hu2023survey} show improvements in the performance of natural language understanding and generation tasks. However, even with such improvements, knowledge-enhanced LLMs still produce errors as measured using common evaluation metrics (e.g. F1 measures for classification). These evaluation metrics are calculated by measuring the difference between an LLM's output and ground truth as provided in fact checking benchmarks such as LAMA \cite{petroni2019language}, KAMEL \cite{kalo2022kamel}, and FActScore \cite{min2023factscore}. 

The errors reported by these metrics are typically assumed to stem from disagreement about facts. But there is another way in which these differences can arise. \textit{Metalinguistic disagreement} \cite{plunkett2023varieties,plunkett2013disagreement,rudolph2023contested} occurs when people argue about the meaning or use of words rather than about facts or ideas. In contrast, a factual disagreement is about what is actually true in the world. Examples of factual disagreement are debating whether a tomato is healthier than an apple, or debating whether Sarah is taller than John; in contrast, examples of metalinguistic disagreement are arguing whether a tomato should be called a fruit or a vegetable, or arguing about what height qualifies as ``tall" when describing a person.

Consider the following scenario: a knowledge-enhanced LLM generates an output that contradicts ground truth provided by a KG. This is used as evidence that the LLM has committed a factual error in its output. However, in producing its output, the knowledge-enhanced LLM has provided a rationale that indicates that there is a disagreement about the meaning of a term that has led to the output. Can this occur in practice? Our hypothesis is that it does. 

Why would this matter? Factual disagreements can be resolved through knowledge graph refinement \cite{paulheim2017knowledge} or through few-shot in-context learning that provides the correct facts to the LLM; however, metalinguistic disagreements may require ontology engineering to address representational issues with a knowledge graph, or the engineering of prompts that incorporate intensional definitions in natural language of concepts for an LLM \cite{allen2023conceptual}. Data governance \cite{khatri2010designing} also acknowledges the importance of establishing metalinguistic agreement of intensional definitions of concepts and relations in natural language and their realization in databases and database schemas; for example, the FAIR principles \cite{wilkinson2016fair,vogt2024fair} specifically urge clear documentation of metadata aligning natural language concepts and metadata in scientific data resources. We therefore argue that distinguishing factual from metalinguistic disagreement between LLMs and KGs is relevant to the practice of knowledge graph and ontology engineering.

\section{Evidence for the occurrence of metalinguistic disagreement in LLMs}
To test our hypothesis that metalinguistic disagreement is a detectable phenomenon, we conducted a simple experiment by fact checking a set of knowledge graph triples aligned with natural language text using an LLM, and then estimating the rate at which metalinguistic disagreement occurs when the LLM determines the triple is not true.

We randomly sampled 100 Wikipedia abstracts from the 10,000 document sample provided in the T-REx dataset, a dataset of large scale alignments between Wikipedia abstracts and Wikidata triples. T-REx has been widely used in the evaluation of LLM-based fact checking and extraction for knowledge graphs \cite{elsahar2018t}. From the total set of triples aligned with the documents in that sample, we then sampled 250 triples. We then defined a zero-shot chain-of-thought classifier \cite{kojima2022large} to assign a truth value to an aligned triple, providing the Wikidata abstract with which it is aligned as context in the LLM prompt \cite{allen2024evaluating}. The classifier was executed to obtain a rationale and a truth value for each of the 250 sampled triples and aligned abstracts, and each result was then processed by a second zero-shot chain-of-thought classifier (using gpt-4o-2024-05-13) acting as an LLM-as-a-judge \cite{chiang2023can}, to classify whether the truth-value-assigning classifier's rationale indicated a metalinguistic disagreement. Processing required a total of 2 inference API calls per alignment, per LLM. Evaluations whose statistics are reported below were conducted during the period from 1 July 2024 to 8 July 2024. Costs incurred through calls to language model APIs totalled less than \$100 USD. Code and data used in the experiments are available in a Github repository\footnote{\url{https://github.com/bradleypallen/trex-metalinguistic-disagreement}}.

\begin{table}[!ht]
\centering
\begin{tabular}{l|p{1cm}|p{1cm}|p{2.5cm}|p{1cm}}
\textbf{LLM} & \textbf{FN} & \textbf{FNR} & \textbf{metalinguistic disagreements} & \textbf{MDR}\\ 
\hline
gpt-4o-2024-05-13 &	26	& 0.104 & 10 & 0.040 \\
gpt-4-0125-preview & 33 & 0.132 & 16 & 0.064 \\
claude-3-haiku-20240307 & 42 & 0.168 & 11 & 0.044 \\
claude-3-opus-20240229 & 51 & 0.204 & 14 & 0.056 \\
claude-3-5-sonnet-20240620 & 52 & 0.208 & 16 & 0.064 \\
meta-llama/Meta-Llama-3-70b-Instruct & 71 & 0.284 & 28 & 0.112 \\
gpt-3.5-turbo & 75 & 0.300 & 29 & 0.116 \\
mistralai/Mixtral-8x7B-Instruct-v0.1 & 78 & 0.312 & 29 & 0.116 \\
mistralai/Mistral-7B-Instruct-v0.3 & 126 & 0.504 & 66 & 0.264 \\
\end{tabular}
\vspace*{2mm}
\caption{Performance by LLM reporting the number of false negatives (FN); the false negative rate (FNR); and the number and rate of metalinguistic disagreements (MDR) detected by the LLM-as-a-judge.}
\label{tab:llm-performance}
\end{table}

As shown in Table \ref{tab:llm-performance}, over the 9 LLMs evaluated, false negative rates over the 250 sampled T-REx triples ranged between 0.104 and 0.504 with a mean of 0.246, and the rate of metalinguistic disagreements between the classifier and Wikidata (i.e., the number of detected metalinguistic disagreements divided by the number of evaluated alignments) ranged between 0.04 and 0.264 with a mean of 0.097. To illustrate the nature of disagreements detected, Table \ref{tab:metalinguistic-disagreement-examples} shows two examples of false negatives from the experiment which exhibit metalinguistic disagreement. The complete set of rationales and classifications is available in the Github repository.

\begin{table}[!ht]
\footnotesize
\centering
\begin{tabular}{p{3cm}|p{11cm}}
\textbf{Wikidata triple} & \textbf{rationale for negative finding} \\ 
\hline
James II of England (Q126188) followed by (P156) Mary II of England (Q130812) & Based on the information provided in the hypothetical world, the knowledge graph triple ``Subject: James II of England, Predicate: followed by, Object: Mary II of England" is false. The passage states that after James II was deposed in 1688, he was replaced by his daughter Mary II, ruling jointly with her husband William III. This means that Mary II did not immediately follow James II, but rather replaced him as the monarch. The predicate ``followed by" implies that Mary II succeeded James II directly, which is not the case. \\
% \hline
% United States of America (Q30) contains the administrative territorial entity (P150) Connecticut (Q779) & The given knowledge graph triple states that the United States of America contains the administrative territorial entity of Connecticut. However, in the hypothetical world where the information about Al Capp is true, this statement is false. The information provided indicates that Al Capp, the creator of the comic strip "Li'l Abner," was from Connecticut. This suggests that Connecticut is a state within the United States, and not a direct subdivision or administrative territorial entity of the United States. Therefore, the statement that the United States of America contains Connecticut as a direct subdivision is false. \\
\hline
chocolate (Q195) made from material (P186) sugar (Q11002) & Based on the information provided, the knowledge graph triple ``Subject: chocolate, Predicate: made from material, Object: sugar" is false. The passage clearly states that chocolate is made from the seeds of the cacao tree, specifically Theobroma cacao, and that it is roasted and ground to produce the chocolate. There is no mention of sugar being a material that chocolate is made from. In fact, the passage states that sugar is added to chocolate to create sweet chocolate, but it is not a primary ingredient that chocolate is made from. Therefore, the given knowledge graph triple is false.
\end{tabular}
\vspace*{2mm}
\caption{Two examples of rationales (generated using claude-3-haiku-20240307) exhibiting metalinguistic disagreement. In both cases, the disagreement is around the meaning of the predicate. For the triple with James II of England as the subject, the LLM states that the historical facts presented in the context are better represented using ``replaced by" (P1366) than ``followed by" (P156); for the triple with chocolate as a subject, the LLM states that ``made from material" (P186) is not an appropriate relation when the object is not a primary ingredient.}
\label{tab:metalinguistic-disagreement-examples}
\end{table}

\section{Proposed benchmark}

We argue that the above results suggest that that metalinguistic disagreement between knowledge graphs and LLMs can occur during fact-checking tasks. However, there are some significant shortcomings in the above approach:
\begin{itemize}
    \item \textbf{Lack of human validation.} The detection of metalinguistic disagreement relies on using an LLM-as-a-judge, which may not be a reliable substitute for human judgment \cite{bavaresco2024llms,thakur2024judging}. This introduces the possibility that the detected ``disagreements" are artifacts of how different LLMs process and generate language, rather than true metalinguistic disagreements. Human review at scale is needed to validate the results. Without this, it's difficult to determine if what the LLMs identify as metalinguistic disagreements align with human judgments.
   \item \textbf{Possible conflation with other error types.} What's interpreted as metalinguistic disagreement could potentially be other types of errors or inconsistencies in LLM outputs, such as hallucinations or context misinterpretations.
    \item \textbf{Limited sample size.} The experiment uses a relatively small sample of 250 triples. A larger-scale study is needed to draw more robust conclusions.
\end{itemize}

We argue that by creating a benchmark metalinguistic disagreement detection dataset that addresses these limitations, we could more confidently assess the occurrence and nature of metalinguistic disagreements in LLM-based fact-checking. This would provide a stronger foundation for investigating our hypothesis and advancing our understanding of how LLMs interpret and disagree about meaning in knowledge graph engineering contexts. 

Specific requirements for such a benchmark include:
\begin{itemize}
    \item \textbf{Human-annotated examples.} A set of fact-checking instances annotated by human experts to identify clear cases of metalinguistic disagreement, factual disagreement, and agreement. This would serve as a gold standard for evaluation.
    \item \textbf{Inter-annotator agreement metrics.} Support the evaluation of system performance using inter-annotator agreement metrics that incorporate knowledge graph ground truth and human annotations to measure the degrees of inter-agent factual and metalinguistic agreement.
    \item \textbf{Multiple knowledge graph sources.} Use triples from different knowledge graphs spanning multiple knowledge domains to account for variations in how relations and concepts are defined across sources, and to test if metalinguistic disagreements are more prevalent in certain areas.
    \item \textbf{Contextual information.} Provide relevant context for each fact-checking instance, similar to the Wikipedia abstracts used by T-REx.
    \item \textbf{Examples with ambiguity, temporal aspects, and gradable predicates.} Deliberately include examples with potential for ambiguity or multiple interpretations to probe the boundaries of metalinguistic disagreement, examples where the truth value of a statement might change over time, to explore how temporal context affects metalinguistic understanding, and examples with gradable predicates (e.g., "tall," "fast") that might be more prone to metalinguistic disagreement.
    \item \textbf{Negative examples.} Include clear cases where no metalinguistic disagreement should occur, to test for false positives.
\end{itemize}

As an initial next step towards this objective, we plan to extend the dataset used in the initial experiments described above in a manner similar to that used in the design and implementation of the SHROOM hallucination detection benchmark \cite{mickus-etal-2024-semeval,allen2024shroom}, through crowdsourcing to incorporate human annotation and increasing the size of the sample of knowledge alignments from the T-REx dataset. Human annotators will be presented with a summary of a Wikipedia page and a statement generated from the Wikidata knowledge graph triple for each alignment, and the annotator must indicate if they disagree with the statement, and if so, whether they disagree on the factuality of the statement or the meaning of any of the terms used in the statement.

In conclusion, we anticipate that such a benchmark can not only shed light on the nature and frequency of metalinguistic disagreements between LLMs and KGs, but also contribute to the ongoing debate about LLMs' capacity for generating meaningful statements. Some have argued that LLMs are incapable of understanding meaning in the way humans do \cite{bender2020climbing}. Others are exploring ways in which LLMs might be capable of at least some limited or partial forms of meaning as a consequence of either the model's pre-training or its grounding through in-context learning \cite{mandelkern2024language,lederman2024language,levinstein2024still,baggio2024referential,grindrod2024large}. We believe that the proposed benchmark can contribute to a more nuanced view of the epistemic status of LLMs relative to KGs based on two-component semantics \cite{berto2022topics,hawke2018theories,hawke2024truth}, and support experimental work in determining whether or not LLMs can generate meaningful statements or be claimed to have beliefs \cite{herrmann2024standards,harding2023operationalising}.

\section*{Acknowledgements}

This work was partially supported by EU’s Horizon Europe research and innovation programme within the ENEXA project (grant Agreement no. 101070305). The authors wish to thank Frank van Harmelen, Levin Hornischer, Filip Ilievski, Jan-Christoph Kalo, Aybüke Özgün, Lise Stork, and Klim Zaporojets for discussions and suggestions that have been invaluable in refining this work.

%%
%% Define the bibliography file to be used
\bibliography{paper}

% \appendix

\end{document}